# LLMs for automatic annotation of Mandarin narrative transcripts

**Qingwen Zhao[1]*, Hongao Zhu[2]*, Yunqi He[3], Rui Wang[1], Aijun Huang[1,4 #], Hai Hu[5#]**

[1]Shanghai Jiao Tong University, China [2]University of California San Diego, USA
[3]The Hong Kong Polytechnic University, China
[4] National Research Center for Language and Well-Being, China
[5]City University of Hong Kong, China
Correspondence: ajhuang@sjtu.edu.cn; hu.hai@outlook.com

## Abstract

Linguistic annotation of transcribed speech is essential for research in language acquisition, language disorders, and sociolinguistics, yet remains labor-intensive and time-consuming. While Large Language Models (LLMs) have shown promise in automating annotation tasks, their ability to handle complex discourse-level annotation in non-English languages remains understudied. This study evaluates whether LLMs can reliably annotate narrative macrostructure—the hierarchical organization of story grammar elements—in spoken Mandarin, using the Multilingual Assessment Instrument for Narratives (MAIN) as a testbed. We compared four LLMs against trained human annotators on narratives produced by children, young adults, and older adults. The best-performing model achieved agreement with human raters ($\kappa = .794$) approaching human-human reliability levels ($\kappa = .872$) while reducing annotation time by 65%, whereas the locally deployable lightweight model performed substantially worse. Annotation difficulty varied systematically by macrostructure element type, with categories requiring subtle semantic differentiation posing persistent challenges. Furthermore, model reliability decreased on young adult narratives, which exhibited greater lexical variation, semantic ambiguity, and multi-element integration within single utterances. These findings suggest that LLMs can effectively support discourse-level annotation in non-English spoken corpora, while highlighting the continued need for human oversight in semantically complex tasks. Our prompt templates are open sourced for future use.

## 1 Introduction

Annotation of conversational discourse is a fundamental step in linguistic research, enabling researchers to identify critical linguistic features for quantitative analysis (Garside et al. 1997; Leech 1997; Nikolaus et al. 2024). However, annotating transcripts of spoken corpora presents greater challenges than annotating written text. Spoken language is characterized by fragmentation, with idea units strung together with minimal connectives, leaving coherence relations implicit and requiring inference (Chafe 1982).

[1] *First two authors contributed equally to this work

Recent advances in Large Language Models (LLMs) have provided researchers with powerful tools for automating linguistic annotation. Studies have shown that ChatGPT outperforms crowd-workers on annotation tasks while costing approximately thirty times less (Gilardi et al. 2023). LLMs can reliably annotate various linguistic phenomena: rhetorical structures in research articles (Kim and Lu 2024), pragmatic features of apologies (Yu et al. 2024), syntactic patterns (Morin and Larsson 2025), and named entities in clinical data (Frei and Kramer 2023; Goel et al. 2023).

Despite these promising findings, several limitations in existing research warrant attention. First, prior studies have predominantly focused on written text, which typically exhibits well-formed grammatical structures. Spoken transcripts, by contrast, often contain incomplete utterances, repetitions, self-corrections, fillers, and non-standard syntactic structures (MacWhinney 2000). Yet only a few studies have examined LLM performance on spoken narrative transcripts (Nikolaus et al. 2024; Ostyakova et al. 2023).

While existing work has addressed discourse-level annotation, the tasks involved are relatively straightforward, typically requiring single-dimension classification such as categorizing rhetorical moves or identifying named entities. Recent work has demonstrated that ChatGPT can achieve human-like performance on speech function annotation in casual conversations (Ostyakova et al. 2023). However, their annotation categories classified individual utterances independently without requiring coherence across the discourse structure. More complex annotation schemes involving multiple interrelated elements across hierarchical narrative structures remain underexplored.

In addition, previous research has treated corpora as homogeneous, without considering whether characteristics of the text producers affect annotation reliability. Children typically produce narratives with simpler syntax and more transparent expression, while adults tend to use more complex syntactic structures and greater lexical diversity (Nippold et al. 2014; Zhang et al. 2019). Whether such variation in linguistic complexity influences LLM annotation performance remains an open question. Finally, existing studies have focused almost exclusively on English, leaving LLM annotation of other typologically distinct languages, such as Mandarin, largely unexplored.

To address these gaps, this study investigates whether state-of-the-art LLMs can reliably annotate narrative macrostructure in spoken Chinese. We use transcripts of oral narratives elicited through the Multilingual Assessment Instrument for Narratives (MAIN, Gagarina et al. 2019a; Luo et al. 2020). It is a standardized tool for assessing narrative skills across languages and age groups (Lindgren et al. 2023). It requires identifying seventeen hierarchically organized story grammar elements (e.g., goals, attempts, outcomes) across multiple causally linked episodes (Gagarina et al. 2019b; Stein and Glenn 1979). This task demands that LLMs maintain long-range coherence, track causal and temporal relations across episodes, and determine whether semantically similar utterances refer to the same or distinct story events. Furthermore, by utilizing narratives produced by children, young adults, and older adults, we examine whether the varying linguistic complexity and coherence associated with different life stages influence LLM performance.

We evaluate four LLMs against gold-standard human annotations: three large-scale models (DeepSeek-R1 [R1], DeepSeek-V3 [V3], and Qwen3-max-preview [Qwen3]) and one locally deployable open-source model (DeepSeek-R1-Distill-Qwen-14B [Qwen14B]). Reliability is assessed using Cohen's Kappa to ensure a robust measure of inter-rater agreement. We also analyze the time efficiency and cost-effectiveness of these models. Our findings reveal that while LLMs can serve as effective first-pass annotators with substantial time savings, their performance varies considerably across models, story grammar elements, and speaker age groups. These results have implications not only for the practical automation of MAIN annotation but also for understanding the current capabilities and limitations of reasoning LLMs in handling complex linguistic annotation tasks.

Our investigation addresses the following research questions:

1. How reliably can LLMs perform discourse-level annotation tasks, specifically macrostructure identification in narratives? Can locally deployable smaller models like Qwen14B achieve comparable reliability to large-scale models?
2. Which types of story grammar elements are more difficult to annotate and exhibit lower agreement for LLMs?
3 Do LLMs exhibit differential annotation performance across narrative texts from distinct age cohorts (children, young and older adults)?

## 2 Method

### 2.1 Macrostructure annotation scheme

Macrostructure is a well-established measure for evaluating narrative discourse organization and has been widely adopted in narrative studies (Altman et al. 2026; Justice et al. 2010; Liles et al. 1995). It refers to the overall organization of narrative discourse and consists of hierarchically arranged story grammar components, including character (the main and supporting characters), setting (time and place), initiating event (an external or internal trigger that prompts character action), goal (the protagonist's objective), attempt (the protagonist's action), outcome (the consequence of that action), and internal response (the protagonist's internal state following the outcome) (Gagarina et al. 2019b; Merritt and Liles 1987; Stein and Glenn 1979).

The narratives analyzed in this study were elicited using MAIN (Gagarina et al. 2019a). MAIN includes four parallel stories, each depicted in a sequence of six pictures that participants are asked to narrate.[2] For example, the Dog story unfolds across three episodes: in the first, a dog chases a mouse but bumps into a tree; in the second, a boy notices his balloon stuck in the tree and retrieves it; in the third, the dog notices the sausages in the boy's bag and steals them while the boy is distracted. Originally designed for children aged 3 to 10 years, recent studies have demonstrated its applicability to adolescents and adults (Gagarina et al. 2019b).

[2] Due to copyright restrictions, the MAIN pictures cannot be reproduced here. The full set of MAIN materials can be found on the MAIN website: https://main.leibniz-zas.de and are accessible after registration.

Two features make MAIN particularly well-suited for evaluating automated annotation. First, its applicability across age groups means that transcripts vary in length, complexity, and linguistic features, providing a rigorous test for model generalization. Second, MAIN provides a fine-grained annotation scheme for macrostructure.

According to the MAIN scoring rubric, macrostructure is evaluated using story structure scores. Story structure quantifies participants' production of story grammar elements (Gagarina et al. 2019a). The maximum story structure score of 17 comprises 2 points for setting elements (time and place) and 15 points for structural components distributed across three episodes. Each episode contains five elements: internal state terms (IST) as initiating event, goal, attempt, outcome, and IST as reaction. All story grammar elements and corresponding response examples for the Dog story are listed in Table 1.

Table 1. Scoring rubric for story structure.

| Story grammar elements | Abbr | Examples of correct responses | Score |
|---|---|---|---|
| Time | T | Time reference, e.g. once upon a time/ one day/ long ago... | 0 1 |
| Location | L | Place reference, e.g. in a forest/ in a park/ in a meadow/ in a field/ by a tree/ near a tree/ by the road | 0 1 |
| Episode 1: Dog (Episode characters: dog and mouse) | | | |
| IST as initiating event 1 | I1 | Dog was playful/ curious; Dog saw a mouse | 0 1 |
| Goal 1 | G1 | Dog wanted to catch/ get/ chase the mouse/ play with the mouse | 0 1 |
| Attempt 1 | A1 | Dog jumped forward/ up; Dog chased/ started to chase | 0 1 |
| Outcome 1 | O1 | Dog bumped his head/ bumped into the tree/ did not get the mouse/ was not quick enough;<br>Mouse escaped/ ran behind the tree/ mouse was too quick | 0 1 |
| IST as reaction 1 | R1 | Dog was disappointed/ angry/ hurt; Mouse was happy/ glad/ relieved | 0 1 |
| Episode 2: Boy (Episode character: boy) | | | |
| IST as initiating event 2 | I2 | Boy was sad/ unhappy/ worried about his balloon;<br>Boy saw the balloon in the tree | 0 1 |
| Goal 2 | G2 | Boy decided/ wanted to get his balloon back | 0 1 |
| Attempt 2 | A2 | Boy was/is pulling/ tried to pull the balloon down from the tree;<br>Boy jumped after the balloon/ reached for (the balloon)/ was/is climbing (the tree) | 0 1 |
| Outcome 2 | O2 | Boy got his balloon back/ again; Balloon was saved | 0 1 |
| IST as reaction 2 | R2 | Boy was glad/ happy/ satisfied/ pleased/ relieved (to get/have his balloon back) | 0 1 |
| Episode 3: Dog (Episode character: dog) | | | |
| IST as initiating event 3 | I3 | Dog saw/ noticed the sausages (in the bag);<br>Dog was hungry/ curious/ keen on the sausages | 0 1 |
| Goal 3 | G3 | Dog wanted/ decided to get/ grab/ eat/ have/ steal the sausages | 0 1 |
| Attempt 3 | A3 | Dog was/is grabbing/pulling/ taking/ stealing the sausages;<br>Dog grabs/pulls/takes the sausages (out of the bag)/ reached for the sausages | 0 1 |
| Outcome 3 | O3 | Dog ate/ got the sausages | 0 1 |
| IST as reaction 3 | R3 | Dog was satisfied/ glad/ pleased/ happy/ not hungry (anymore) | 0 1 |
| | | Total score | 17 |

**Note.** Adapted from Gagarina et al. (2019a).

Manual annotation of story structure scores poses significant challenges in both time efficiency and consistency. Even for trained annotators, annotating a single story takes approximately 10 minutes, making large-scale annotation highly time-consuming. Beyond time investment, the scoring criteria require subtle semantic distinctions. As illustrated in Table 1, within Episode 3, the Goal (G3), Attempt (A3), and Outcome (O3) all involve the same core elements: the dog, grab/eat, and the sausages. The distinction lies in aspectual and temporal nuances: G3 captures the dog's intention (e.g., "wanted to eat"), A3 describes the action in progress (e.g., "was grabbing"), and O3 marks the completed event (e.g., "ate/got the sausages"). Annotators must carefully distinguish between prospective, ongoing, and completed actions, a task that demands both semantic sophistication and attention to delicate linguistic cues.

## 2.2 Data source

The transcripts utilized in this study were drawn from a larger corpus (Authors in prep), comprising narrative samples from 207 native Mandarin-speaking participants. The sample spans a wide developmental range from early childhood to late adulthood (ages 3–88), comprising 116 children (ages 3–7), 20 young adults (mean age 25), and 71 older adults (ages 60–90) (cf. Table 2). Each participant produced two different stories, resulting in a total pool of 414 narratives.

Table 2. Demographic characteristics of the participants

| Group | Age (y) | N | Mean Age (y;m) | Age Range |
|---|---|---|---|---|
| | 3 | 23 | 3;6 | 3;2 - 3;11 |
| | 4 | 25 | 4;4 | 4;0 - 4;11 |
| **children** | 5 | 26 | 5;4 | 5;0 - 5;11 |
| | 6 | 22 | 6;4 | 6;0 - 6;11 |
| | 7 | 20 | 7;4 | 7;0 - 7;11 |
| **young adults** | 20 | 20 | 25;1 | 21;10 - 31;2 |
| | 60 | 23 | 64;10 | 60;2 - 69;11 |
| **older adults** | 70 | 24 | 75;9 | 70;3 - 79;11 |
| | 80 | 24 | 83;10 | 80;4 - 88;2 |
| **Total** | | | 207 | |

We employed proportional stratified sampling to select 15% from each age group, yielding 64 narratives (18 children, 3 young adults, 11 older adults). All transcripts were prepared in the CHAT format (MacWhinney 2000), segmented into T-units (Hunt 1965). We retained all spoken language features without preprocessing, including incomplete utterances, repetitions, and self-corrections. These samples were annotated independently by two trained human raters and four LLMs to enable systematic comparison of inter-rater agreement.

### 2.3 Expert and automated annotation of macrostructure elements

Two trained human annotators first jointly tagged 8 stories outside the testing corpus to establish reliability. They then independently scored 64 stories. For each of the 17 possible elements, annotators judged the presence or absence of each element. Inter-annotator reliability was assessed using Cohen's Kappa (Cohen 1960).

We employed four LLMs: three API-based models, DeepSeek-R1 (R1), DeepSeek-V3 (V3), and Qwen3-max-preview (Qwen3), and one locally deployable model, DeepSeek-R1-Distill-Qwen-14B (Qwen14B).

The annotation prompt was developed by child language acquisition experts and refined through iterative pilot testing. The final version achieved the highest accuracy for both stories (see Appendix A and B). Consistent with human annotation criteria, LLMs were evaluated on their judgments regarding the presence or absence of each macrostructure element. Agreement between each LLM and human annotators was assessed using Cohen's $\kappa$. Specifically, kappa was calculated separately for each LLM-human pair, and the mean of the two $\kappa$ values (i.e., LLM vs. Annotator 1 and LLM vs. Annotator 2) was reported as the human-LLM agreement. Human consensus served as the gold standard.

## 3 Results

### 3.1 Model performance, efficiency, and cost comparison

Table 3 presents the inter-rater agreements among human raters and LLMs. According to Landis and Koch's (1977) criteria, human annotators achieved almost perfect agreement ($\kappa$ = .872). Notable performance differences emerged across LLMs.

Among the four LLMs, R1 demonstrated the best performance, achieving substantial agreement with human raters ($\kappa$ = .794). V3 ($\kappa$ = .751) and Qwen3($\kappa$ = .725) also indicated substantial agreement. The locally deployable model Qwen14B exhibited the lowest agreement ($\kappa$ = .530 ), representing only moderate agreement.

Several patterns emerged from these results. First, larger-scale LLMs demonstrated substantially higher agreement with human annotators than their lightweight counterparts, with $\kappa$ values ranging from .725 to .794 compared to .530 for Qwen14B. Among the three large-parameter models, both DeepSeek models (R1 and V3) outperformed Qwen3. Second, within the DeepSeek family, the reasoning model (R1) marginally outperformed its non-reasoning counterpart (V3) by .043 in $\kappa$ value.

Table 3. Inter-rater agreements among human raters and LLMs (Cohen's $\kappa$)

| Comparison Pair | reasoning model | κ | interpretation |
|---|---|---|---|
| **inter-human-annotator** | | 0.872 | almost perfect |
| **human vs. DeepSeek-R1** | √ | 0.794 | substantial |
| **human vs. DeepSeek-V3** | | 0.751 | substantial |
| **human vs. Qwen3-max-preview** | √ | 0.725 | substantial |
| **human vs. DeepSeek-R1-Distill-Qwen-14B** | √ | 0.53 | moderate |

Beyond annotation accuracy, time and cost efficiency are critical considerations for practical implementation (cf. Table 4). The three API-based models annotated all 64 narrative texts in 10 to 30 minutes at minimal cost (R1: ¥0.66; V3: ¥0.08; Qwen3: free during preview), whereas the locally deployed Qwen14B required approximately 38 hours with a GPU cost of ¥50.61. Given R1's superior performance, we used it to annotate all 414 narrative texts in the corpus, completing the task in approximately 3 hours at a cost under ¥5. Human raters then verified each file, averaging only 3 minutes per file (approximately 21 hours total). Compared to the estimated 69 hours for complete manual annotation, this LLM-assisted workflow required only 24 hours, a 65% reduction in annotation time.

Table 4. Cost of compute, API or GPU renting

| Models | tokens | cost (￥) | estimated time (min) |
|---|---|---|---|
| **DeepSeek-R1** | 320,393 | 0.66 (API) | 30 |
| **DeepSeek-V3** | 112,575 | 0.08 (API) | 10 |
| **Qwen3-max-preview** | 131,065 | free trial (API) | 10 |
| **DeepSeek-R1-Distill-Qwen-14B** | 635,726 | 50.61 (GPU)<br>1.32 * 38 hours = 50.61 | 2280 |

The substantial performance gap between API-based and locally deployable models suggest that smaller open-source models are currently inadequate for complex semantic annotation tasks like MAIN scoring. Consequently, the 14B model was excluded from further detailed analysis to focus on the performance patterns of more capable systems. The broader implications of this model choice are discussed in Section 4.1.

### 3.2 Element-specific agreement patterns

To examine the performance of large language models in narrative analysis, we categorized story grammar elements based on their annotation difficulty (cf. Table 5). We identified a clear distinction between elements that demonstrate high reliability and those that act as annotation bottlenecks.

The high-performance group showed robust model-human alignment and can be further divided into two tiers. Time (mean $\kappa = .932$) and outcome (mean $\kappa = .814$) achieved almost perfect agreement, with all models maintaining high stability ($\kappa \geq .718$). Meanwhile, location (mean $\kappa = .767$) and IST as reaction (mean $\kappa = .728$) reached substantial agreement. Within this group, R1 showed unexpected difficulty with location ($\kappa = .559$), performing substantially below other compute-intensive models ($\kappa \geq .826$) and falling into the moderate range, which suggests model-specific variation even in generally reliable categories.

The low-performance group, in contrast, highlights the current inferential boundaries of LLMs. Elements in this category achieved only moderate agreement, including attempt (mean $\kappa = .450$), IST as initiating event (mean $\kappa = .503$), and goal (mean $\kappa = .606$). Notably, attempt emerged as the most significant bottleneck across all models; no model reached the .70 threshold for substantial agreement, with values ranging from .308 to .630. The locally deployable Qwen14B struggled significantly with these semantically complex categories, failing to reach even the moderate threshold ($\kappa < .41$) for goal, attempt, and IST as initiating event. In particular, Qwen14B's poor performance on goal ($\kappa = .280$) acted as a significant outlier that dragged down the category's mean to a moderate level ($\kappa = .606$); in contrast, all other LLMs demonstrated much stronger performance on this element, all reaching substantial agreement levels between .672 and .791. This pattern suggests that while smaller models can handle lexically explicit information, they lack the deep inferential capacity required for discourse-level semantic interpretation.

Table 5. Inter-rater agreement by story grammar category (Cohen's $\kappa$)

| Story grammar elements | Human-human | Human-R1 | Human-V3 | Human-Qwen3 | Human-Qwen14B | Model mean | Interpretation |
|---|---|---|---|---|---|---|---|
| **Time** | 1 | 0.966 | 0.965 | 0.929 | 0.867 | 0.932 | almost perfect |
| **Outcome** | 0.888 | 0.857 | 0.829 | 0.851 | 0.718 | 0.814 | almost perfect |
| **Location** | 0.905 | 0.559 | 0.826 | 0.89 | 0.791 | 0.767 | substantial |
| **IST as reaction** | 0.924 | 0.774 | 0.844 | 0.801 | 0.493 | 0.728 | substantial |
| **Goal** | 0.853 | 0.791 | 0.672 | 0.68 | 0.28 | 0.606 | moderate |
| **IST as initiating event** | 0.747 | 0.751 | 0.573 | 0.377 | 0.309 | 0.503 | moderate |
| **Attempt** | 0.73 | 0.63 | 0.454 | 0.407 | 0.308 | 0.45 | moderate |

**Note.** Cohen's Kappa interpretation: < 0.41 (fair); 0.41-0.60 (moderate); 0.61-0.80 (substantial); 0.81-1.00 (almost perfect).

### 3.3 Variation in model performance across age groups

Table 6 presents Cohen's Kappa coefficients measuring agreement between human raters and four LLMs across three age groups. Several key findings emerge from this analysis.

Table 6. Inter-rater agreement across age groups (Cohen's $\kappa$)

| Group | Human-human | Human-R1 | Human-V3 | Human-Qwen3 | Human-Qwen14B | Model mean | Interpretation |
|---|---|---|---|---|---|---|---|
| **chi (Children)** | 0.869 | 0.773 | 0.782 | 0.748 | 0.613 | 0.729 | substantial |
| **eld (Elderly)** | 0.865 | 0.846 | 0.721 | 0.691 | 0.405 | 0.666 | substantial |
| **you (Young)** | 0.909 | 0.686 | 0.624 | 0.674 | 0.368 | 0.588 | moderate |

**Note.** Cohen's Kappa interpretation: < 0.41 (fair); 0.41-0.60 (moderate); 0.61-0.80 (substantial); 0.81-1.00 (almost perfect).

Across all three age groups, human-human agreement remained consistently high, with kappa values exceeding 0.85. This indicates that the annotation guidelines were well-defined and that human raters maintained uniform standards across different speaker populations. Notably, human raters achieved their highest agreement on young adults' narratives ($\kappa = .909$).

However, model-human agreement displayed the opposite pattern. The mean model-human agreement scores show a clear gradient: models performed best on children's narratives ($\kappa = .729$, substantial agreement), followed by elderly speakers ($\kappa = .666$, substantial agreement), and performed worst on young adults' narratives ($\kappa = .588$, moderate agreement). This divergence between human and model performance is noteworthy: the narratives that were easiest for human raters to annotate consistently proved most challenging for LLMs.

## 4 Discussion

### 4.1 LLM performance in macrostructure annotation

This study investigated whether state-of-the-art LLMs can reliably annotate narrative macrostructure in spoken Mandarin corpora, using the MAIN assessment framework as a testbed for evaluating complex discourse-level semantic annotation. Our findings demonstrate that LLMs can achieve agreement levels approaching human performance, though systematic limitations persist.

Regarding annotation accuracy, R1 emerged as the best-performing model, achieving a .794 agreement with human annotators, approaching the human-human agreement of .872. This finding extends previous work on LLM-based linguistic annotation, which has primarily focused on written English text and relatively constrained annotation tasks (Goel et al. 2023; Kim and Lu 2024; Morin and Larsson 2025). Our results suggest that contemporary LLMs can handle substantially more complex annotation challenges, including those requiring semantic interpretation across fragmented spoken utterances in non-Indo-European languages.

Notably, R1's performance ($\kappa = .794$) is comparable to human inter-rater reliability in prior research. Using the same MAIN framework with Mandarin-speaking children, Sheng et al. (2020) reported human-human agreement ranging from .843 to .864. This suggests that the best-performing LLM can approximate trained human coders in narrative macrostructure annotation.

However, smaller locally deployable models showed limited capability for this task. As reported in Section 3.1, Qwen14B achieved only moderate agreement with human annotators while requiring longer processing time and higher costs. This suggests that complex semantic annotation tasks like MAIN scoring currently exceed the capabilities of smaller open-source models. For researchers prioritizing data privacy, locally deployable models may serve as a starting point, but outputs require more extensive human verification. This creates a practical trade-off: higher accuracy necessitates API-based access, which involves transmitting data to external servers.

Beyond accuracy, LLM-assisted annotation offers substantial efficiency gains. Using R1 for initial tagging followed by human verification reduced total annotation time by approximately 65%, with negligible API costs. Our findings suggest that an AI-assisted, human-verified workflow could substantially accelerate large-scale data collection efforts without sacrificing reliability.

### 4.2 The challenge of tense-aspect processing and relational inference

Having established overall model performance, we now turn to a more detailed analysis of the drivers behind human-model disagreement. Human and LLM annotations for a representative child narrative transcript are detailed in Table 7. Our analysis reveals that while LLMs excel at identifying elements with overt linguistic markers, they struggle with categories requiring fine-grained semantic boundaries and pragmatic inference.

Table 7. Sample narrative transcripts of story one with macrostructure annotations

| ID | Sentence | Human 1 | Human 2 | R1 | V3 | Qwen3 | Qwen14B |
|---|---|---|---|---|---|---|---|
| 1 | *有一天小狗出来玩。*'One day the little dog came out to play' | T, I1 | T, I1 | T | T | T | T |
| 2 | *它发现一只老鼠。*'It found a mouse.' | I1 | I1 | I1 | I1 | I1 | I1 |
| 3 | *想吃它* (It) wanted to eat it | G1 | G1 | G1 | G1 | G1 | G1 |
| 4 | *可是老鼠跑进了<洞 (.) 里> [//] 树洞里。*'But the mouse ran into <the hole (.)> [//] the tree hole.' | O1 | O1 | O1 | O1 | O1 | O1 |
| 5 | *它找不着。*'It couldn't find (it).' | O1 | Null | O1 | O1 | O1 | O1 |
| 6 | *只好用头伸进 (.) 去看。*'(It) had to stick its head in (.) to look.' | Null | Null | Null | A1 | A1 | A1 |
| 7 | 旁边的小孩拎着一堆肉肠。'The child nearby was carrying a bunch of sausages.' | Null | Null | Null | Null | I3 | I3 |
| 8 | *发现了一个气球。*'(He) found a balloon.' | I2 | I2 | Null | I2 | I2 | I2 |
| 9 | *哦<他> [/] 他就去找气* 'Oh <he> [/] he went to get the bal(loon).' | Null | Null | A2 | Null | G2 | G2 |
| 10 | *他就想去够气球。*'He wanted to reach for the balloon.' | G2 | G2 | G2 | G2 | G2, A2 | A2 |
| 11 | *<小> [/] 小狗缠着想吃那个肉肠。*'<The little> [/] The little dog pestered (him) wanting to eat those sausages.' | G3 | G3 | I3, G3 | I3, G3 | I3, G3, A3 | G3 |
| 12 | *<后> [/] 后来小朋友够到气球了。*'<Later> [/] Later the child reached the balloon.' | O2 | O2 | O2 | A2, O2 | O2 | O2 |
| 13 | *可是我们小狗<在> [/] 在开心地吃着肉肠。*‘But our little dog <was> [/] was happily eating the sausages.’ | O3, R3 | O3, R3 | O3, R3 | A3, O3 | O3, R3 | O3 |
| 14 | *最后 &-uh 小朋友 &-uh 很开心。*‘Finally &-uh the child &-uh was very happy.’ | R2 | R2 | R2 | R2 | R2 | R2 |
| 15 | *小狗也很开心。*‘The little dog was also very happy.’ | R3 | R3 | R3 | R3 | R3 | R3 |

**Note.** Story grammar element abbreviations: T = Time; L = Location; I = Internal state term as initiating event; G = Goal; A = Attempt; O = Outcome; R = Internal state term as reaction. Numbers 1–3 indicate episode number.
Transcription conventions: [/] = repetition; [//] = correction; [///] = reformulation; (.) = silent pause; &-uh = filled pause.

#### 4.2.1 Successes: lexical explicitness and grammatical anchors

Four elements including Time, Location, Outcome, and IST as Reaction demonstrated consistently high agreement. This success reflects the presence of unambiguous lexical cues: Time and Location were anchored by explicit temporal/prepositional markers. Outcomes were signaled by the Mandarin perfective marker *le* (*了*) and resultative verb complements (e.g., *跑进了* ‘ran into’). Similarly, IST as Reaction achieved high consistency because emotional vocabulary (e.g., *很开心* ‘very happy’) belongs to an explicit lexical class requiring minimal contextual inference. This suggests that for surface-level linguistic features, LLMs have achieved human-like reliability.

#### 4.2.2 Challenge 1: aspectual sensitivity and the action-verb bias

The identification of Goal and Attempt emerged as a significant bottleneck, primarily due to an action-verb bias. Despite clear prompt instructions to distinguish Goal (intention) from Attempt (action), models frequently allowed specific verbs to override their grammatical environment. This is evident in the treatment of the verb *gòu* (*够* ‘reach’) across different contexts. In Sentence 10 (*他就想去够* ‘He wanted to reach...’), models like Qwen3 erroneously assigned an Attempt tag, ignoring the volitional auxiliary *xiǎng* (*想* ‘want’) which marks intention. Conversely, in Sentence 12 (*够到气球了* ‘reached the balloon’), where the resultative complement *dào* (*到*) and perfective *le* (*了*) explicitly signal completion, V3 still added an Attempt tag alongside the Outcome.

These overlapping annotations suggest that LLMs prioritize lexical-level verb matching over higher-level syntactic embedding and aspectual marking. This tendency to overlook the logical boundaries between intention, attempt, and completion points to the potential benefit of more sophisticated prompting. For instance, incorporating contrastive negative examples could help models better navigate these delicate semantic boundaries.

#### 4.2.3 Challenge 2: relational inference and the keyword trap

The moderate agreement on IST as Initiating Event stems from the models' inability to distinguish between the presence of an object and a character's perception of that object. This element requires a confirmed link, such as seeing or noticing, between a character and a stimulus that prompts character action (Stein and Glenn, 1979). When such perceptual relation was marked by explicit perceptual verbs like *fā xiàn* (*发现* 'found/noticed') in Sentences 2 and 8, models correctly identified the initiating event.

However, in the absence of an explicit perceptual link, models frequently fell into a keyword trap, defaulting to object-keyword matching regardless of the character's perspective. In Sentence 7 ('the child nearby was carrying sausages'), Qwen3 and Qwen14B assigned an IST tag simply because "sausages" appeared, ignoring the narrative fact that the protagonist (dog) had not yet perceived them. This is further evidenced in Sentence 11 ('the dog pestered him wanting to eat sausages'), where R1, V3 and Qwen3 correctly identified a Goal but simultaneously assigned an IST tag due to the mere co-occurrence of "dog" and "sausages." This suggests that LLMs struggle to inhibit keyword-driven associations even when they conflict with the actual narrative function of the utterance.

Given these observations, a tiered approach to annotation may be more reliable. While entity-based elements like Time and Location appear suitable for automation, relational elements involving character perspectives likely still require human oversight to ensure accuracy.

### 4.3 The impact of linguistic complexity on model performance across age groups

The quantitative results in Section 3.3 revealed that LLMs achieved highest agreement on children's narratives, with performance declining notably on young adult data. We propose three explanations for this finding.

First, young adults employed more varied and complex vocabulary, which may confuse LLMs. Table 8 presents annotations for IST as initiating event 3, which the coding manual specifies as "Dog saw/noticed the sausages." The child participant (chi-56) produced "*结果小狗看到了袋子里的香肠* 'As a result, the little dog saw the sausage in the bag'", closely mirroring this definition. All four LLMs achieved perfect agreement with human raters. In contrast, young adult participant you-5 used *wén dào* (*闻到* 'smelled') rather than *kàn dào* (*看到* 'saw'), representing a sensory modality shift. Although this conveys the same narrative function, Qwen14B misannotated it as Outcome 2, suggesting that even minor lexical deviations from prototypical expressions can disrupt model annotation.

Second, young adults' lexical choices introduced greater semantic ambiguity. Participant you-1 produced "*这时候没有捉到老鼠的小狗盯上了那袋烧鸡* 'At this moment, the little dog who didn't catch the mouse set its eyes on that bag of roast chicken'". The verb *dīng shàng* (*盯上* 'set eyes on') carries semantic weight beyond neutral perception, implying intentionality or desire. This may explain why R1 and Qwen14B annotated both IST as Initiating Event 3 and Goal 3, while human raters identified only the initiating event.

Table 8. Human and LLM annotations for IST as Initiating Event 3 across age groups

| Particpant ID | Sentence | Human 1 | Human 2 | R1 | V3 | Qwen3 | Qwen14B |
|---|---|---|---|---|---|---|---|
| **chi-56** | *结果小狗看到了袋子里的香肠。*‘As a result, the little dog saw the sausage in the bag.’ | I3 | I3 | I3 | I3 | I3 | I3 |
| **you-5** | *小狗就闻到了香肠的气味。* ‘The little dog smelled the scent of the sausage.’ | I3 | I3 | I3 | I3 | I3 | O2 |
| **you-1** | *这时候没有捉到老鼠的小狗盯上了那袋烧鸡。*‘At this moment, the little dog who didn’t catch the mouse set its eyes on that bag of roast chicken.’ | I3 | I3 | I3, G3 | I3 | I3, G3 | O2, I3 |

**Note.** Story grammar element abbreviations: I = Internal state term as initiating event; G = Goal; O = Outcome; Numbers 1–3 indicate episode number. chi = children, you = young adults, eld = elderly

Third, young adults tended to integrate multiple story grammar elements within single utterances. Table 9 illustrates this pattern. For IST as Reaction 3, child participant chi-112 produced “*小狗也很开心* ‘The little dog was also very happy’”, encoding only the emotional reaction. All models correctly annotated this single-element sentence. However, young adult participant you-5 produced “*<小狗在后面> [/] 小狗在后面高兴地吃到了它的这个香肠* ‘The little dog in the back happily ate its sausage’”, which integrates location (*在后面* ‘in the back’), manner/emotion (*高兴地* ‘happily’), and the completed action (*吃到了* ‘ate’) within one clause. Although human raters consistently identified Outcome 3 and IST as Reaction 3, both R1 and Qwen14B exhibited over-annotation. R1 added Attempt 3, while Qwen14B produced the most extensive over-annotation, additionally marking IST as Initiating Event 3, Goal 3, and Attempt 3.

Table 9. Human and LLM annotations for IST as Reaction 3 across age groups

| Participant ID | Sentence | human 1 | human 2 | R1 | V3 | Qwen3 | Qwen14B |
|---|---|---|---|---|---|---|---|
| **chi-112** | *小狗也很开心。*‘The little dog was also very happy.’ | I3 | I3 | I3 | I3 | I3 | I3 |
| **You-5** | *<小狗在后面> [/] 小狗在后面高兴地吃到了它的这个香肠。*‘<The little dog in the back> [/] The little dog in the back happily ate its sausage.’ | O3, I3 | O3, I3 | A3, O3, I3 | O3, I3 | O3, I3 | I3, G3, A3, O3, I3 |

**Note.** Story grammar element abbreviations: I = Internal state term as initiating event; G = Goal; A = Attempt; O = Outcome; Numbers 1–3 indicate episode number.

These findings suggest that superior model performance on children's narratives stems not from simpler story content, but from linguistic transparency. Children tend to map one utterance to one story grammar element using straightforward, prototypical expressions, whereas adult speakers' varied vocabulary, semantic ambiguity, and syntactically integrated constructions create annotation challenges that current LLMs have yet to reliably overcome.

### 4.4 Implications, limitations, and future directions

Our findings carry several implications for using LLMs in linguistic research. First, LLMs can serve as effective first-pass annotators for complex discourse-level tasks, provided human verification remains part of the workflow. The 65% time savings represent meaningful efficiency gains for labor-intensive projects. Second, model selection matters: larger models consistently outperformed lightweight counterparts, and reasoning capabilities did not guarantee superior performance. Researchers should conduct pilot testing rather than assuming newer models perform better. Third, annotation difficulty is element-specific. Tasks involving lexically explicit categories (like Outcomes or temporal markers) suit LLM automation, while those requiring fine-grained semantic differentiation (like distinguishing Attempts from Goals) demand substantial human oversight. Fourth, narratives with greater lexical diversity and syntactic integration require more intensive human verification, while corpora with simpler utterance-to-element mappings permit greater reliance on automation.

Several limitations warrant acknowledgment. Although we utilized the standardized MAIN protocol, this study is localized to a single task in Mandarin. Future research should extend this paradigm to diverse narrative tasks and other languages to determine if the observed aspectual biases and keyword traps are universal. Furthermore, our prompts primarily relied on positive examples. Future studies should incorporate explicit negative examples to help models navigate delicate semantic boundaries in complex relational inference tasks.

## 5 Conclusion

This study provides the first systematic evaluation of LLM performance on narrative macrostructure annotation in spoken Mandarin, using the widely adopted MAIN framework as a testbed. Our findings demonstrate that state-of-the-art LLMs can achieve near-human reliability on this complex task, offering substantial practical benefits for research scalability. At the same time, systematic limitations persist: models perform well on elements with explicit lexical markers but struggle with those requiring tense-aspect processing, pragmatic inference, or fine-grained semantic boundary detection. Performance also varies across speaker populations, with linguistically complex adult narratives posing greater challenges than children's more transparent productions. Lightweight models remain substantially less reliable than their compute-intensive counterparts. These findings contribute both to the practical implementation of LLM-assisted annotation workflows and to our theoretical understanding of current LLMs' capabilities and limitations in handling the discourse-level semantic challenges that characterize naturalistic language data.

## Appendices

### Appendix A: Prompt for Story One

### Original Chinese version

现在要对儿童、青年、老人的叙事文本进行标注。他们的叙事文本是基于故事图片，故事图片包括三个故事情节：

**情节一：狗追老鼠**

小狗看到一只老鼠，想要去追老鼠。小狗往前一跳，结果没追到老鼠，反而撞到了树上。狗很生气，老鼠则逃走了。

**情节二：男孩找气球**

男孩因分心看狗追老鼠，不慎将气球飞上树梢。男孩因为他的气球有些难过，他决定取回气球，于是爬上来树去拿，最后成功拿回了气球，男孩很高兴。

**情节三：小狗偷香肠**

趁男孩爬树取气球时，小狗看到了男孩袋子里的香肠，狗想去吃香肠，于是它去扒开袋子，吃到了香肠。

我把被试的叙事文本发给你，请你帮我标注叙事文本中是否提到了 A0-A16 这 17 个事件。每个事件的关键词如下表，如果被试提到了对应的正确事件，就给分：

- A0 表示整个故事的背景时间，关键词：很久以前、有一天、从前;
- A1 表示整个故事的背景地点，关键词：在森林里、在公园、在草地、在路边;
- A2 表示情节一的引发事件，关键词：狗在玩、狗很好奇、狗看到一只老鼠;
- A3 表示情节一的目标，关键词：狗想抓老鼠、狗想得到老鼠、狗想追赶蝴蝶、狗想和老鼠一起玩；或者其他意思相近的表达：狗准备抓老鼠。表示想要完成，但还没有付诸行动。
- A4 表示情节一的尝试，关键词：狗向前跳、狗向前扑过去; 表示实际做出的尝试或行动。
- A5 表示情节一的结果，关键词：狗撞到头、狗没有抓到老鼠、狗与老鼠一起玩、老鼠逃走了、老鼠跑到树后面、老鼠跑太快了;
- A6 表示情节一的内部反应，关键词：狗很难过、狗很生气、狗很受伤、老鼠很开心、老鼠很高兴;
- A7 表示情节二的引发事件，关键词：男孩为他的气球而伤心、男孩为他的气球而难过、男孩为他的气球而担心、男孩看到他的气球在树上;
- A8 表示情节二的目标，关键词：男孩决定找回他的气球、男孩想找回他的气球; 或者其他意思相近的表达：准备去拿气球。表示想要完成，但还没有付诸行动。
- A9 表示情节二的尝试，关键词：男孩尝试拉他的球、男孩跳上树去拉气球; 表示实际做出的尝试或行动。
- A10 表示情节二的结果，关键词：男孩拿回了他的气球、男孩的气球得救了;

- A11 表示情节二的内部反应，关键词：男孩很开心、男孩很高兴、男孩很满足;
- A12 表示情节三的引发事件，关键词：狗注意到香肠、狗看见香肠、狗很饿、狗很好奇; 或者其他意思相近的表达：狗看着香肠口水直流。
- A13 表示情节三的目标，关键词：狗想去拿香肠、狗决定去拿香肠、狗想去吃香肠、狗决定去吃香肠、狗想去偷香肠、狗决定去偷香肠; 或者其他意思相近的表达：狗准备去拿香肠。表示想要完成，但还没有付诸行动。
- A14 表示情节三的尝试，关键词：狗去拿香肠、狗去接近香肠、狗去拿袋子、狗去接近袋子; 表示实际做出的尝试或行动。
- A15 表示情节三的结果，狗吃到香肠、狗得到香肠;
- A16 表示情节三的内部反应，关键词：狗很满足、狗很高兴、狗很饱;

打分时请注意：

1. 不一定需要完整地提到关键词的所有表述，只要意思相近，但是省略了主语或者宾语，或者用第一人称、第二人称或者其他近义词表述也可以，比如"男孩决定找回他的气球"的相似表达可以是"他准备去拿气球"、"小孩想把气球拿下来"。
2. 被试的叙述顺序可能不一定完全按照三个情节的展开顺序，而是会跳跃，导致一句话可能包含多个事件，比如在"小男孩把气球够下来以后 ，小狗已经吃完一根香肠了"这句话中，包含了 A10 与 A15 两个事件，横跨了情节二和情节三，因此需要同时标注 A10，A15。
3. 如果被试产出的故事情节偏离的设定的情节，则不用给错误的话语进行标注，比如假设被试说"小男孩准备拿香肠去换钱"，这一情节在原故事图片中没有提及，则直接忽略这句话，不用打分。
4. 区分"目标"和"尝试"，"目标"表示想要完成，但还没有付诸行动，用将来时表示，而"尝试"则表示实际做出的尝试或行动，用进行时或者完成体表示。比如，"小狗就跑过去吃"，这句话是现在进行时，表明这是"尝试"而非"目标"。
5. A1-A6 与情节一对应，A7-A11 与情节二对应，A12-A16 与情节三对应，请勿错位。

以字典形式直接返回"位置字典"：包含不同事件在对话中的行号，事件可能存在于单个/多个语句;若事件未提及返回 Null。下面是一个例子：

示例文本=
""""
1 有一天有一个小老鼠
2 它想回洞里
3 然后小狗它在这里抓这个小老鼠
4 有一个小朋友走过来
5 拿着香肠和气球
6 小老鼠躲到了树后面

7 它撞到了上面
8 他的气球飞了
9 气球挂到了树上
10 他的东西放在了地上
11 小狗看见想去吃
12 他摘下气球
13 发现他的已经被小狗吃完了
”””

示例位置字典={‘A0’: [1], ‘A1’: Null, ‘A2’: Null, ‘A3’: Null, ‘A4’: [3], ‘A5’: [6, 7], ‘A6’: Null, ‘A7’: [8, 9], ‘A8’: Null, ‘A9’: Null, ‘A10’: [12], ‘A11’: Null, ‘A12’: [11], ‘A13’: [11], ‘A14’: Null, ‘A15’: [13], ‘A16’: Null}

**Translated English version**

You are now required to annotate narrative texts produced by children, young adults and older adults. Their narratives are based on story pictures containing three episodes:
**Episode 1: Dog Chases Mouse**
A dog sees a mouse and wants to chase it. The dog jumps forward but fails to catch the mouse and instead bumps into a tree. The dog feels angry, and the mouse escapes.
**Episode 2: Boy Retrieves Balloon**
A boy becomes distracted watching the dog chase the mouse and accidentally lets his balloon float up into a tree. The boy feels sad about his balloon and decides to retrieve it, so he climbs the tree to get it. He successfully retrieves the balloon and feels happy.
**Episode 3: Dog Steals Sausage**
While the boy is climbing the tree to retrieve his balloon, the dog notices sausages in the boy’s bag. The dog wants to eat the sausages, so it opens the bag and eats them.

I will provide you with participants’ narrative texts. Please help me annotate whether the following 17 events (A0-A16) are mentioned in the narrative. The keywords for each event are listed below. If the participant mentions the corresponding event correctly, mark it:

- **A0** represents the temporal setting of the entire story. Keywords: once upon a time / one day / long ago
- **A1** represents the spatial setting of the entire story. Keywords: in a forest / in a park / in a meadow / by the road / near a tree
- **A2** represents the initiating event of Episode 1 (IST). Keywords: dog was playful / curious; dog saw a mouse
- **A3** represents the goal of Episode 1. Keywords: dog wanted to catch / get / chase the mouse / play with the mouse; or similar expressions: dog was preparing to catch the mouse. This indicates intention without action yet taken.
- **A4** represents the attempt in Episode 1. Keywords: dog jumped forward / up; dog chased / started to chase. This indicates actual attempts or actions taken.

- **A5** represents the outcome of Episode 1. Keywords: dog bumped his head / bumped into the tree / did not get the mouse / mouse was not quick enough; mouse escaped / ran behind the tree / mouse was too quick
- **A6** represents the internal reaction in Episode 1 (IST). Keywords: dog was disappointed / angry / hurt; mouse was happy / glad / relieved
- **A7** represents the initiating event of Episode 2 (IST). Keywords: boy was sad / unhappy / worried about his balloon; boy saw the balloon in the tree
- **A8** represents the goal of Episode 2. Keywords: boy decided / wanted to get his balloon back; or similar expressions: preparing to get the balloon. This indicates intention without action yet taken.
- **A9** represents the attempt in Episode 2. Keywords: boy was/is pulling / tried to pull the balloon down from the tree; boy jumped after the balloon / reached for the balloon / was/is climbing the tree. This indicates actual attempts or actions taken.
- **A10** represents the outcome of Episode 2. Keywords: boy got his balloon back / again; balloon was saved
- **A11** represents the internal reaction in Episode 2 (IST). Keywords: boy was glad / happy / satisfied / pleased / relieved
- **A12** represents the initiating event of Episode 3 (IST). Keywords: dog saw / noticed the sausages (in the bag); dog was hungry / curious / keen on the sausages; or similar expressions: dog was drooling over the sausages
- **A13** represents the goal of Episode 3. Keywords: dog wanted / decided to get / grab / eat / have / steal the sausages; or similar expressions: dog was preparing to get the sausages. This indicates intention without action yet taken.
- **A14** represents the attempt in Episode 3. Keywords: dog was/is grabbing / pulling / taking / stealing the sausages; dog grabs/pulls/takes the sausages (out of the bag) / reached for the sausages; dog tried to + VERB (get, take). This indicates actual attempts or actions taken.
- **A15** represents the outcome of Episode 3. Keywords: dog ate / got the sausages
- **A16** represents the internal reaction in Episode 3 (IST). Keywords: dog was satisfied / glad / pleased / happy / not hungry

**Important annotation guidelines:**

(1) Partial matches are acceptable. The participant doesn’t need to mention all elements of a keyword phrase. Similar meanings are acceptable even if subjects, objects are omitted, or first/second person pronouns or synonyms are used. For example, “boy decided to get his balloon back” can be expressed as “he was preparing to get the balloon” or “the child wanted to take down the balloon.”

(2) Participants may not follow the chronological order of the three episodes and may jump between them. A single sentence may contain multiple events. For example, in the sentence “After the boy got the balloon down, the dog had already finished eating a sausage,” both A10 and A15 are mentioned, spanning Episodes 2 and 3. Therefore, both A10 and A15 should be annotated.

(3) If the participant's narrative deviates from the preset storyline, do not annotate incorrect utterances. For example, if the participant says "the boy was preparing to exchange the sausages for money," this plot is not mentioned in the original story pictures, so ignore this sentence and do not score it.
(4) Distinguish between "Goal" and "Attempt." "Goal" indicates intention without action yet taken, expressed in future tense, while "Attempt" indicates actual attempts or actions taken, expressed in progressive or perfect aspect. For example, "the dog ran over to eat" is in present progressive tense, indicating this is an "Attempt" rather than a "Goal."
(5) A1-A6 correspond to Episode 1, A7-A11 correspond to Episode 2, and A12-A16 correspond to Episode 3. Do not misalign them.
Return the **"Position Dictionary"** in dictionary format: containing the line numbers where different events appear in the dialogue. Events may exist in single/multiple utterances; if an event is not mentioned, return Null. Below is an example:
**Example text:**
1 One day there was a little mouse
2 It wanted to go back to its hole
3 Then the dog was here catching this little mouse
4 A child walked over
5 Holding sausages and a balloon
6 The little mouse hid behind the tree
7 It bumped into the top
8 His balloon flew away
9 The balloon got stuck in the tree
10 He put his things on the ground
11 The dog saw it and wanted to eat
12 He picked the balloon
13 Found that his had already been eaten by the dog
**Example Position Dictionary:**
{'A0': [1], 'A1': Null, 'A2': Null, 'A3': Null, 'A4': [3], 'A5': [6, 7], 'A6': Null, 'A7': [8, 9], 'A8': Null, 'A9': Null, 'A10': [12], 'A11': Null, 'A12': [11], 'A13': [11], 'A14': Null, 'A15': [13], 'A16': Null}

**Appendix B: Prompt for Story Two**

**Original Chinese version**

现在要对儿童、青年、老人的叙事文本进行标注。他们的叙事文本是基于故事图片，故事图片包括三个故事情节：
**情节一、小猫捉蝴蝶：**
小猫看到一只蝴蝶，想到抓蝴蝶，于是扑过去，结果没抓到蝴蝶，反而跌进草丛受伤生气。
**情节二、男孩捞球：**
男孩因分心看猫追蝴蝶，不慎将球掉入水中，男孩因为他的球有些难过，他决定捞回他的球，于是用鱼竿捞球，最后成功拿回了球，男孩很高兴。
**情节三、小猫偷鱼：**
趁男孩捞球时，小猫看到了男孩鱼桶里的鱼，猫想去吃鱼，于是去偷鱼，成功吃到了鱼。

我把被试的叙事文本发给你，请你帮我标注叙事文本中是否提到了 A0-A16 这 17 个事件。每个事件的关键词如下表，如果被试提到了对应的正确事件，就给分：

- A0 表示整个故事的背景时间，关键词：很久以前、有一天、从前;
- A1 表示整个故事的背景地点，关键词：在森林里、在湖边、在河边、在草丛
- A2 表示情节一的引发事件，关键词：猫在玩、猫很好奇、猫看到一只蝴蝶;
- A3 表示情节一的目标，关键词：猫想抓蝴蝶、猫想得到蝴蝶、猫想追赶蝴蝶、猫想和蝴蝶一起玩; 或者其他意思相近的表达：猫准备抓蝴蝶。表示想要完成，但还没有付诸行动。
- A4 表示情节一的尝试，关键词：猫向前跳、猫向前扑过去; 表示实际做出的尝试或行动。
- A5 表示情节一的结果，关键词：猫摔倒在小树里、猫没有抓到蝴蝶、猫与蝴蝶一起玩、蝴蝶逃走了、蝴蝶飞走了、蝴蝶飞太快了
- A6 表示情节一的内部反应，关键词：猫很难过、猫很生气、猫很受伤、蝴蝶很开心、蝴蝶很高兴;
- A7 表示情节二的引发事件，关键词：男孩为他的球而伤心、男孩为他的球而难过、男孩为他的球而担心、男孩看到他的球在水里;
- A8 表示情节二的目标，关键词：男孩决定找回他的球、男孩想找回他的球；或者其他意思相近的表达：我想要捡回我的球、他想捡球、他准备捡球。表示想要完成，但还没有付诸行动。
- A9 表示情节二的尝试，关键词：男孩捞他的球; 表示实际做出的尝试或行动。
- A10 表示情节二的结果，关键词：男孩拿回了他的球、男孩的球得救了;
- A11 表示情节二的内部反应，关键词：男孩很开心、男孩很高兴、男孩很满足;
- A12 表示情节三的引发事件，关键词：猫注意到鱼、猫看见鱼、猫很饿、猫很好奇;

- A13 表示情节三的目标，关键词：猫想去拿鱼、猫决定去拿鱼、猫想去吃鱼、猫决定去吃鱼、猫想去偷鱼、猫决定去偷鱼；或者其他意思相近的表达：它想过去吃、它想过去吃、它准备去吃鱼。表示想要完成，但还没有付诸行动。
- A14 表示情节三的尝试，关键词：猫去拿鱼、猫去接近鱼；或者其他意思相近的表达：它就去拿了。表示实际做出的尝试或行动。
- A15 表示情节三的结果，关键词：猫吃到鱼、猫得到鱼;
- A16 表示情节三的内部反应，关键词：猫很满足、猫很高兴、猫很饱;

打分时请注意：
（1）不一定需要完整地提到关键词的所有表述，只要意思相近，但是省略了主语或者宾语，或者用第一人称、第二人称或者其他近义词表述也可以，比如"男孩决定找回他的球"的相近表达可以是"他决定找回球"、"我决定找回我的球"、"小孩决定找回来"，这些表述也可以给分。
（2）被试的叙述顺序可能不一定完全按照三个情节的展开顺序，而是会跳跃，导致一句话可能包含多个事件，比如在"小男孩把球捞上来以后 , 小猫也吃到鱼了"这句话中，包含了 A10 与 A15 两个事件，横跨了情节二和情节三，因此需要同时标注 A10，A15
（3）如果被试产出的故事情节偏离的设定的情节，则不用给错误的话语进行标注，比如假设被试说"小男孩准备拿鱼去卖钱"，这一情节在原故事图片中没有提及，则直接忽略这句话，不用打分。
（4）区分"目标"和"尝试"，"目标"表示想要完成，但还没有付诸行动，用将来时表示，而"尝试"则表示实际做出的尝试或行动，用进行时或者完成体表示。比如，"小猫就跑过去吃"，这句话是现在进行时，表明这是"尝试"而非"目标"。
（5）A1-A6 与情节一对应，A7-A11 与情节二对应，A12-A16 与情节三对应，请勿错位。

以字典形式直接返回"位置字典"：包含不同事件在对话中的行号，事件可能存在于单个/多个语句;若事件未提及返回 Null。下面是一个例子：

示例文本=
""""""
1 有一天,小猫看见一只黄色的蝴蝶
2 它想扑上去抓住它
3 这时有一个开朗的男孩过来
4 不一会儿蝴蝶就飞走了
5 它掉到了草丛里
6 它感觉又痛又生气
7 他出来后
8 那个男孩看见他很痛
9 手松开
10 球掉了
11 小猫这时看见了鱼

12 他想过去吃
13 这时男孩他的球掉到了水里
14 他用鱼竿把球拿了上来
15 他没注意到后面的小猫
16 小猫吃了他里面的一条鱼
”””

示例位置字典={‘A0’: [3], ‘A1’: Null, ‘A2’: [3], ‘A3’: [4], ‘A4’: Null, ‘A5’: [6, 7], ‘A6’: [8], ‘A7’: [12, 15], ‘A8’: Null, ‘A9’: Null, ‘A10’: [16], ‘A11’: Null, ‘A12’: [13], ‘A13’: [14], ‘A14’: Null, ‘A15’: [18], ‘A16’: Null}

**Translated English version**

You are now required to annotate narrative texts produced by children, young adults, and elderly participants. Their narratives are based on story pictures containing three episodes:

**Episode 1: Cat Catches Butterfly**

A cat sees a butterfly and wants to catch it, so it pounces. However, it fails to catch the butterfly and instead falls into the bushes, getting hurt and angry.

**Episode 2: Boy Retrieves Ball**

A boy becomes distracted watching the cat chase the butterfly and accidentally drops his ball into the water. The boy feels sad about his ball and decides to retrieve it, so he uses a fishing rod to get the ball. He successfully retrieves the ball and feels happy.

**Episode 3: Cat Steals Fish**

While the boy is retrieving his ball, the cat notices fish in the boy’s bucket. The cat wants to eat the fish, so it steals the fish and successfully eats it.

I will provide you with participants’ narrative texts. Please help me annotate whether the following 17 events (A0-A16) are mentioned in the narrative. The keywords for each event are listed below. If the participant mentions the corresponding event correctly, mark it:

- **A0** represents the temporal setting of the entire story. Keywords: once upon a time / one day / long ago
- **A1** represents the spatial setting of the entire story. Keywords: in the forest / by the lake / at the river bank / by the water / by the shore / in a meadow / in the bushes
- **A2** represents the initiating event of Episode 1 (IST). Keywords: cat was playful / curious; cat saw a butterfly
- **A3** represents the goal of Episode 1. Keywords: cat wanted to catch / get / chase the butterfly / play with the butterfly; or similar expressions: cat was preparing to catch the butterfly. This indicates intention without action yet taken.
- **A4** represents the attempt in Episode 1. Keywords: cat jumped forward / up; cat chased / started to chase; cat tried to + VERB (catch, get, grab, take). This indicates actual attempts or actions taken.

- **A5** represents the outcome of Episode 1. Keywords: cat fell into the bush / did not get the butterfly / was not quick enough; butterfly escaped / flew away / was too quick
- **A6** represents the internal reaction in Episode 1 (IST). Keywords: cat was disappointed / angry / hurt; butterfly was happy / glad
- **A7** represents the initiating event of Episode 2 (IST). Keywords: boy was sad / unhappy / worried about his ball; boy saw the ball in the water
- **A8** represents the goal of Episode 2. Keywords: boy decided / wanted to get his ball back; or similar expressions: I want to pick up my ball / he wants to pick up the ball / he is preparing to pick up the ball. This indicates intention without action yet taken.
- **A9** represents the attempt in Episode 2. Keywords: boy was/is pulling / tried to pull the ball out of the water. This indicates actual attempts or actions taken.
- **A10** represents the outcome of Episode 2. Keywords: boy got his ball back / again; the ball was saved
- **A11** represents the internal reaction in Episode 2 (IST). Keywords: boy was glad / happy / pleased / satisfied / relieved (to get/have his ball back)
- **A12** represents the initiating event of Episode 3 (IST). Keywords: cat was hungry / curious / keen on the fish; cat noticed / saw the fish
- **A13** represents the goal of Episode 3. Keywords: cat wanted / decided to get / grab / eat / have / steal the fish; or similar expressions: it wanted to go eat / it was preparing to eat the fish. This indicates intention without action yet taken.
- **A14** represents the attempt in Episode 3. Keywords: cat was/is grabbing / pulling / taking / stealing the fish; cat grabs/pulls/takes the fish (out of the bucket) / reached for the fish; cat tried to + VERB (get, take); or similar expressions: it went to take it. This indicates actual attempts or actions taken.
- **A15** represents the outcome of Episode 3. Keywords: cat ate / got the fish
- **A16** represents the internal reaction in Episode 3 (IST). Keywords: cat was satisfied / glad / pleased / happy / not hungry (any more

**Important annotation guidelines:**

(1) Partial matches are acceptable. The participant doesn't need to mention all elements of a keyword phrase. Similar meanings are acceptable even if subjects, objects are omitted, or first/second person pronouns or synonyms are used. For example, "boy decided to get his ball back" can be expressed as "he decided to get the ball back," "I decided to get my ball back," or "the child decided to get it back." These expressions should also be scored.

(2) Participants may not follow the chronological order of the three episodes and may jump between them. A single sentence may contain multiple events. For example, in the sentence "After the boy got the ball up, the cat also ate the fish," both A10 and A15 are mentioned, spanning Episodes 2 and 3. Therefore, both A10 and A15 should be annotated.

(3) If the participant's narrative deviates from the preset storyline, do not annotate incorrect utterances. For example, if the participant says "the boy was preparing to take the fish to sell for money," this plot is not mentioned in the original story pictures, so ignore this sentence and do not score it.

(4) Distinguish between "Goal" and "Attempt." "Goal" indicates intention without action yet taken, expressed in future tense, while "Attempt" indicates actual attempts or actions taken, expressed in progressive or perfect aspect. For example, "the cat ran over to eat" is in present progressive tense, indicating this is an "Attempt" rather than a "Goal."

(5) A1-A6 correspond to Episode 1, A7-A11 correspond to Episode 2, and A12-A16 correspond to Episode 3. Do not misalign them.

Return the **"Position Dictionary"** in dictionary format: containing the line numbers where different events appear in the dialogue. Events may exist in single/multiple utterances; if an event is not mentioned, return Null. Below is an example:

**Example text:**

1 One day, the cat saw a yellow butterfly
2 It wanted to pounce and catch it
3 At this time, a cheerful boy came over
4 Soon the butterfly flew away
5 It fell into the bushes
6 It felt painful and angry
7 After he came out
8 The boy saw he was in pain
9 His hand loosened
10 The ball dropped
11 The cat then saw the fish
12 It wanted to go eat
13 At this time the boy's ball fell into the water
14 He used the fishing rod to get the ball up
15 He didn't notice the cat behind him
16 The cat ate one of the fish inside

**Example Position Dictionary:**

{'A0': [1], 'A1': Null, 'A2': [1], 'A3': [2], 'A4': Null, 'A5': [5], 'A6': [6], 'A7': [10, 13], 'A8': Null, 'A9': [14], 'A10': [14], 'A11': Null, 'A12': [11], 'A13': [12], 'A14': Null, 'A15': [16], 'A16': Null}